\documentclass{article} 
\usepackage[preprint]{colm2026_conference}

\usepackage{microtype}
\usepackage{hyperref}
\usepackage{url}
\usepackage{booktabs}
\usepackage{graphicx}
\usepackage{multirow}
\usepackage{booktabs}
\usepackage{array}
\usepackage{adjustbox}
\usepackage{amsmath}
\usepackage{subcaption}
\usepackage{amssymb}
\usepackage{enumitem}
\usepackage{pifont}
\usepackage[dvipsnames]{xcolor}
\usepackage{float}
\usepackage{wrapfig}
\usepackage[normalem]{ulem}
\useunder{\uline}{\ul}{}
\usepackage[font=small]{caption}

\usepackage{lineno}

\usepackage{color-edits}
\addauthor[Anjali]{anj}{blue}
\addauthor[Oreva]{oreva}{red}
\addauthor[Aarohi]{aarohi}{purple}
\addauthor[Graham]{gn}{magenta}

\newcommand{\aspace}{\hspace{0.5em}}

\definecolor{darkblue}{rgb}{0, 0, 0.5}
\hypersetup{colorlinks=true, citecolor=darkblue, linkcolor=darkblue, urlcolor=darkblue}

\title{\textsc{IdioleX}: Unified and Continuous Representations \\ for Idiolectal and Stylistic Variation}

\author{
\vspace{-1em}\\
\textbf{Anjali Kantharuban}\textsuperscript{1}\aspace
\textbf{Aarohi Srivastava}\textsuperscript{2}\aspace
\textbf{Fahim Faisal}\textsuperscript{3}\aspace
\textbf{Orevaoghene Ahia}\textsuperscript{4} \\
\textbf{Antonios Anastasopoulos}\textsuperscript{3, 5}\aspace
\textbf{David Chiang}\textsuperscript{2}\aspace
\textbf{Yulia Tsvetkov}\textsuperscript{4}\aspace
\textbf{Graham Neubig}\textsuperscript{1} \\
\vspace{-0.2cm} \\
\textsuperscript{1}Carnegie Mellon University\aspace 
\textsuperscript{2}University of Notre Dame\aspace
\textsuperscript{3}George Mason University \\
\textsuperscript{4}University of Washington\aspace
\textsuperscript{5}Archimedes Research Unit \\
\vspace{-0.2cm}\\
\textbf{Correspondence}: \href{mailto:anjaliruban@cmu.edu}{\texttt{anjaliruban@cmu.edu}}
}

\begin{document}

\ifcolmsubmission
\linenumbers
\fi

\maketitle

\begin{abstract}
Existing sentence representations primarily encode \emph{what} a sentence says, rather than \emph{how} it is expressed, even though the latter is important for many applications. 
In contrast, we develop sentence representations that capture style and dialect, decoupled from semantic content. We call this the task of \emph{idiolectal representation learning}.
We introduce \textsc{IdioleX}, a framework for training models that combines supervision from a sentence's provenance with linguistic features of a sentence's content, to learn a continuous representation of each sentence's style and dialect.
We evaluate the approach on dialects of both Arabic and Spanish.
The learned representations capture meaningful variation and transfer across domains for analysis and classification. 
We further explore the use of these representations as training objectives for stylistically aligning language models.
Our results suggest that jointly modeling individual and community-level variation provides a useful perspective for studying idiolect and supports downstream applications requiring sensitivity to stylistic differences, such as developing diverse and accessible LLMs.

\begin{center}
   \textbf{Code:} {\url{github.com/AnjaliRuban/IdioleX}}
\end{center}
\end{abstract}
\section{Introduction}

\begin{wrapfigure}[15]{r}{0.45\textwidth}
    \centering
    \includegraphics[width=\linewidth]{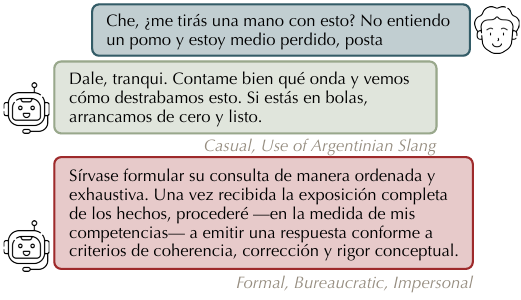}
    \caption{\textsc{IdioleX} used to compare the idiolectal alignment between user input and GPT 5.1 generations in casual Argentinian Spanish.}
    \label{fig:example}
\end{wrapfigure}

Much of mainstream evaluation and optimization of large language models (LLMs) prioritizes semantic correctness \citep{clark-etal-2020-tydi, lewis-etal-2020-mlqa, liu-2023-is, pmlr-v235-chiang24b, singh-etal-2025-global, kim-etal-2025-biggen}.
User-specific linguistic and stylistic adaptation receives comparatively less emphasis in standard benchmarks, despite substantial literature on controllable generation, style transfer, and personalization \citep{jin-etal-2022-deep, dong-etal-2023-steerlm, durandard-etal-2025-language}.
Although semantic correctness is essential, maximizing understanding in languages with diverse speaker populations requires modeling variation across dialect, register, and style, which we collectively refer to as \emph{idiolect} \citep{allan2013common}.

In 1948, Bernard Bloch coined the term \emph{idiolect} to refer to ``the totality of the possible utterances of one speaker at one time in using language to interact with one other speaker'' \citep{bloch1948set}.
In modern linguistics, the concept of idiolect encompasses all the person-, interlocutor-, and context-specific considerations of language use.
Since LLMs are not stable interlocutors with consistent personalities and backgrounds \citep{durandard-etal-2025-language}, the lack of common linguistic ground can result in misunderstanding of both users' and systems' idiolect \citep{frisch-giulianelli-2024-llm, fleisig-etal-2024-linguistic, nias2025culture, lin-etal-2025-assessing}.

Prior NLP work related to idiolect has largely appeared under task-specific formulations such as dialect identification \citep[DID;][]{tiedemann-ljubesic-2012-efficient, lui-cook-2013-classifying} and authorship attribution \citep[AA;][]{stamatatos-etal-1999-automatic, zhu-jurgens-2021-idiosyncratic, rivera-soto-etal-2021-learning, wegmann-etal-2022-author}. 
Although these tasks rely on similar linguistic signals to one another, they have historically been studied in isolation.
Similarly, there have been attempts to learn style-aware representations for authorship verification using either contrastive learning over sentence pairs based on authorship \citep{wegmann-etal-2022-author} or explicit linguistic features \citep{patel-etal-2023-learning}, but none that combine both nor any that make use of more granulated levels of idiolectal proximity.

\textbf{We propose \textsc{IdioleX}, a training framework for learning idiolectal sentence representations with minimal supervision}. 
These representations allow us to capture \emph{idiolectal similarity} between utterances and to support a range of applications, including comparisons between text segments and downstream uses such as post-training or evaluation of LLMs.
\textsc{IdioleX} is language-agnostic and relies on weak supervision from hierarchical proximity and LLM-generated linguistic features rather than task-specific labels. 
The resulting representations are designed to de-prioritize semantic content while encoding stylistic and linguistic structure in a continuous space, enabling applications across analysis, classification, and alignment.

Specifically, we show that \textsc{IdioleX} representations are successful in multiple contexts:
\begin{enumerate}[nosep]
    \item \textbf{A Base for Existing Tasks} that outperforms prior neural approaches on DID and AA across Arabic and Spanish varieties (\S\ref{sec:class}).
    \item \textbf{Embeddings for Similarity Scoring} that encode linguistic structure and dialectal variation with minimal semantic interference~(\S\ref{sec:repval}).
    \item \textbf{A Style Alignment Training Objective} to improve idiolectal and dialectal alignment during post-training (\S\ref{sec:post}).
\end{enumerate}

\section{Idiolectal Representation Learning}

\label{sec:task}

Idiolectal variation -- differences in how individuals use language in terms of lexicon, morphology, syntax, and register -- has long been studied in work on authorship attribution, dialect identification, and style transfer \citep{zhu-jurgens-2021-idiosyncratic, wegmann-etal-2022-author, patel-etal-2025-styledistance}.
We build on these lines of research towards idiolectal representation learning (IRL) to encode stylistic and dialectal properties independent of the task.

\paragraph{Stylistic \& Dialectal Similarity}
\label{sec:tasksim}
We assume that stylistic and dialectal similarity correlate with a \emph{proximity hierarchy}: that language use tends to be most similar within the same individual and same document, followed by the same individual across documents, then among speakers within a shared dialect community, and least similar across different dialect communities.
As such, idiolectal similarity can be modeled as a continuous space reflecting shared linguistic features across speakers and communities. We use this proximity hierarchy (same document $>$ same author $>$ same dialect $>$ cross-dialect) to provide weak supervision during training while avoiding discrete categorization.

\paragraph{Relationship to Semantic Content}
\label{sec:tasksem}
We consider the following desideratum: if two documents are stylistically similar but discuss different topics, their representations should remain relatively close, whereas documents that are semantically similar but differ substantially in style or dialect should be more distant.
While stylistic and semantic signals are intertwined, our objective emphasizes linguistic form over topical content.

\section{\textsc{IdioleX}}

\begin{figure*}[t]
    \centering
    \includegraphics[width=0.9\linewidth]{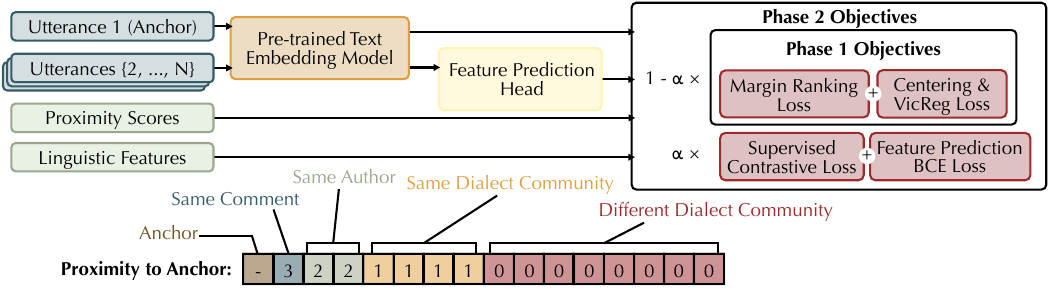}
    \caption{\textsc{IdioleX} training framework. During training, all batches are sampled such that every individual item can act as an anchor for contrastive learning, necessitating that there are $2^{3-n}$ samples for each proximity score $n \in [0,3]$.}
    \label{fig:architecture}
\end{figure*}

\label{sec:feats}

We introduce \textsc{IdioleX}, a framework for training sentence representations that encode idiolectal variation by combining weak supervision from proximity information (defined above, \S\ref{sec:task}) with LLM-extracted linguistic features.
Our goal is to encourage representations that are consistent with the view of stylistic variation described in Section~\ref{sec:task}. 
Rather than relying on a single supervision signal, we integrate multiple sources that provide complementary perspectives on linguistic variation, seen in Figure~\ref{fig:architecture}.

\subsection{Data Schema}
\label{sec:data}

\paragraph{Data Sampling \& Filtering}
We construct training data from publicly available Reddit comments associated with regional communities.
For each language, we identify the main regional subreddits and collect comments from the inception of the subreddit until December 2024 using the Pushshift archive \citep{baumgartner2020pushshift} and filter for language and poor-quality. 
A full list of subreddits and filtering criteria are provided in Appendix~\ref{app:data}.

\paragraph{Proximity Scores}
Because stylistic similarity is not directly observable, we approximate it using relative comparisons derived from authorship and community structure.
For each anchor sentence, we sample comparison sentences representing varying degrees of proximity:
(3) same comment, (2) same author, (1) same dialect community, and (0) different communities.
These groupings provide a weak but structured signal that encourages the representation space to reflect stylistic relationships without requiring explicit labels.
Proximity scores are assigned for each pair of sentences between 0 (different community) and 3 (same comment) based on the most exclusive category the sentence belongs to, as seen in Figure~\ref{fig:architecture}.
This ranking is inherently noisy; however, across a large number of samples, it still provides useful gradients.  
To mitigate this noise, we introduce an additional level of supervision using LLM-generated dialectal features.

\paragraph{Linguistic Features}

In addition to relational supervision, we incorporate automatically extracted linguistic features that reflect common dimensions of dialectal variation \citep{williams2008dialect}.
These linguistic features are extracted from linguistic textbooks on Spanish \citep{SpanishTextbook} and Arabic \citep{ArabicTextbook} dialectology. 
Feature extraction is performed using an LLM prompted to identify whether each attribute is present on a sentence-by-sentence level, since LLMs have been demonstrated to have meta-linguistic capabilities \citep{begus2025large}.
In this work, we use GPT 5 mini\footnote{gpt-5-mini-2025-08-07} for feature extraction.
We view these signals as approximate cues whose limitations (varying extraction quality) are mitigated through complementary supervision from relational structure.
Details of prompts, feature definitions, and annotation procedures are provided in Appendix~\ref{app:feat_extraction}.

\subsection{Architecture}

The \textsc{IdioleX} framework builds on a pretrained sentence encoder that serves as the backbone for representation learning.
This design allows the framework to be applied to different languages by substituting the underlying pretrained model.
Importantly, the architecture is intentionally minimal: it does not assume task-specific structures or discrete label prediction, and can be used as a general-purpose representation learner for stylistic and dialectal variation.

Past work has demonstrated that transformer-based language models encode this information across different layers \citep{tenney2019bert}.
Sentence representations are obtained by applying learned layer-wise attention over all transformer layers (inspired by \cite{rei2020comet}) followed by token-level mean pooling. 
The resulting representations are mean-centered and L2-normalized during training.

Training proceeds in two stages due to limitations on the quantity of data that can be annotated with linguistic features. 
First, the encoder is optimized on the full train set using a relative similarity objective -- a margin ranking loss -- that encourages sentences with higher expected stylistic overlap to be closer in representation space than less related sentences. 
This pre-training primes the model to move from semantic to stylistic encoding and makes use of the significantly larger amount of unannotated data to maximize generalizability.
In the second stage, feature-level supervision is incorporated alongside the margin ranking loss on a smaller subset of the training data through auxiliary objectives that encourage the representation to encode dimensions captured by the extracted linguistic features. 
Combining relational and feature signals allows the model to balance global structure with interpretable variation.
We additionally apply normalization and regularization techniques to stabilize training and reduce representation collapse, promoting a well-conditioned embedding space suitable for downstream analysis.

\paragraph{Margin Ranking Loss}

The margin ranking loss calculates the pairwise difference in the semantic similarity $s$ of each sentence with a given \textit{anchor} sentence, $a$, and then compares it to the difference in proximity score $r$ between sentences $i$ and $j$: 
\begin{equation*}
\begin{split}
    \text{MRL}(a, i, j) = & \max(0, -(r(x_a, x_i) - r(x_a, x_j))  \times (s(x_a, x_i) - s(x_a, x_j)) + \lambda). 
\end{split}
\end{equation*}
In other words, we enforce that higher-proximity pairs have a greater cosine similarity by at least $\lambda$.
Because we structure our mini-batch such that every sentence can act as an anchor, see Figure~\ref{fig:architecture}, we sum over every possible anchor and pair of comparison sentences for the final loss.
During training, the margin $\lambda$ starts at $0$, then increases linearly to $0.5$.

\paragraph{Feature Prediction}
The feature prediction head takes the idiolectal sentence representations and applies a two-layer prediction model for the set of LLM-generated linguistic features, with a binary cross-entropy loss applied on the logits.

\paragraph{Supervised Contrastive Loss} 
The supervised contrastive loss acts as an additional similarity loss, but in place of the ranking from earlier, it uses the Jaccard similarity $w(x_i, x_j)$ over the feature vectors $x$ as the weights for a supervised contrastive loss for sentences in a batch $B$ and $\tau = 0.07$:
\begin{equation*}
        \text{SCL}(i, j) = w(x_i, x_j) \log \frac{e^{\text{sim}(z_i, z_j)/ \tau} }{\displaystyle\sum_{k \in B} e^{\text{sim} (z_i, z_k) / \tau}}.
\end{equation*}
As with MRL, we sum over all combinations of $i, j \in B$. This allows us to better identify the true similarity of sentences from different dialects as a linguist would see them, but we are limited by the chosen features. 
Both feature losses are used after a larger pre-training stage because the features are only available for a subset of the data.

\paragraph{Centering \& Variance Constraints}

We enforce mean centering around 0 using a learned $\mu$ and combine a variance and decorrelation loss to encourage a standard deviation of 1.0 across each dimension and decorrelation between dimensions to avoid anisotropy.
The centering vector $\mu$ is learned jointly with the encoder parameters using batch-wise estimates.

Each language's \textsc{IdioleX} model is trained independently, starting from a language-specific pre-trained base model, AraBert v2\footnote{\url{https://huggingface.co/aubmindlab/bert-base-arabertv2}} for Arabic and Bertin\footnote{\url{https://huggingface.co/bertin-project/bertin-roberta-base-spanish}} for Spanish \citep{antoun2020arabert, delarosa2022bertinefficientpretrainingspanish}. 
Data from 10 authors from each dialect is withheld for each of the development and test sets (150 in total per set).
Hyperparameters, model configurations, and optimization settings are reported in Appendix~\ref{app:train_details} to facilitate reproducibility.
\section{Performance on Classification Benchmarks}
\label{sec:class}

We show that \textsc{IdioleX} provides strong performance on downstream style-related tasks with the addition of task-specific heads.
We evaluate this ability for dialect identification (DID) and authorship attribution (AA).
For each task, we evaluate three versions of the models trained via \textsc{IdioleX}.
\begin{enumerate}[nosep]
    \item \textbf{\textsc{IdioleX}}: The model representations as-is, with a trained classification head, demonstrating the performance of the representations in a cross-domain context.
    \item \textbf{Fine-tuned \textsc{IdioleX}}: The model representations with a trained classification head after finetuning on the train split for each task.
    \item \textbf{Fine-tuned \textsc{IdioleX} + Lexical}: The model representations with a trained classification head after finetuning, ensembled with a trained logistic regression head that is fed TF-IDF representations.
    The \textsc{IdioleX} framework prioritizes the encoding of the morphological and syntactic features associated with idiolect (see Appendix~\ref{app:feat_extraction}), so the addition of explicit lexical features aids in tasks where lexical variation gives significant information.
\end{enumerate}

We evaluate \textsc{IdioleX} against two baselines: 
(1) a fine-tuned BERT-based system (similar training to Fine-tuned \textsc{Idiolex}), and (2) a centroid clustering baseline. 
For any systems based on BERT, we use the same monolingual models used as the pre-trained encoder for training our \textsc{IdioleX} models.
The centroid clustering baseline is constructed with frozen encoder representations from which dialect centroids are calculated on the train data.
Where comparable, we also report winning shared-task scores on the datasets we evaluate on.
More information on classification model training is found in Appendix~\ref{app:class}, and further details regarding the baselines can be found in Appendix~\ref{app:class_baseline}.

\subsection{Dialect Identification}

\begin{figure}
\begin{minipage}[b]{0.6\textwidth}
    \centering
    \small
    \begin{tabular}{lcccc}
    \toprule
    & \multicolumn{2}{c}{\textbf{Spanish}} & \multicolumn{2}{c}{\textbf{Arabic}} \\
    \textbf{Models} & \textbf{F1} & \textbf{EM} & \textbf{F1} & \textbf{EM} \\
    \midrule
    \textsc{IdioleX} & \textbf{0.85} & 0.62 & 0.43  & 0.43 \\
    Fine-tuned \textsc{IdioleX} & 0.84 & \textbf{0.63} & 0.61 & 0.61 \\
    Fine-tuned \textsc{Idiolex} + Lexical & \textbf{0.85} & \textbf{0.63} & 0.66 & 0.66 \\
    \midrule
    Fine-tuned BERT & 0.80 & 0.59 & 0.56 & 0.56 \\
    Centroid Clustering w/ BERT &  0.77& 0.57 &  0.20 & 0.22 \\
    Centroid Clustering w/ E5 & 0.74& 0.54 & 0.15 & 0.17 \\ 
    \midrule
    \citealp{saleva-palen-michel-2024-brandeis} & 0.82 & 0.50 & -- & -- \\
    \citealp{abu-kwaik-saad-2019-arbdialectid} & -- & -- & \textbf{0.67} & \textbf{0.67} \\
    \citealp{samih-etal-2019-qc} & -- & -- & 0.59 & 0.59 \\
    \bottomrule
    \end{tabular}
    \captionof{table}{DID F1-score and exact match accuracy (EM) results on Spanish (DSL-ML) and Arabic (MADAR 26). External baselines are: the top scoring submission to DSL-ML, the top scoring submission to MADAR 26, and the top scoring \textit{neural} submission to MADAR 26.}
    \label{tab:dialectID}
\end{minipage}%
\hfill
\begin{minipage}[b]{0.35\textwidth}
    \centering
    \includegraphics[width=0.85\linewidth]{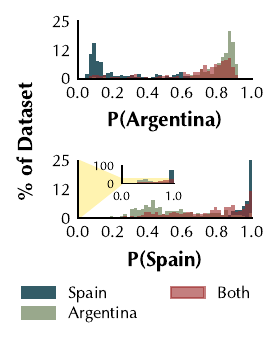}
    \captionof{figure}{Non-fine-tuned \textsc{IdioleX} classification model's likelihood distribution over samples on multi-label Spanish DID.}
    \label{fig:spanish_did}
\end{minipage}
\end{figure}

DID relies heavily on stylistic cues such as lexical choices, morphology, and syntactic patterns, which may overlap across dialects.
For Spanish, we use the multi-label DSL-ML 2024 shared task dataset \citep{chifu-etal-2024-vardial,zampieri-etal-2024-language}. 
For Arabic, we evaluate on MADAR 26 \citep{bouamor-etal-2018-madar,bouamor-etal-2019-madar}, which contains sentences annotated with 25 city-level dialect labels along with Modern Standard Arabic.

Table~\ref{tab:dialectID} reports F1 scores and exact match accuracy (EM) on DSL-ML and MADAR 26.
For DSL-ML, our models outperform both the winning VarDial 2024 submission by \cite{saleva-palen-michel-2024-brandeis} and all baselines. 
Fine-tuning and lexical ensembling provide marginal benefits, demonstrating the strength of \textsc{IdioleX} representations despite the domain mismatch.
Figure~\ref{fig:spanish_did} shows the predicted likelihood distributions from the base \textsc{IdioleX} model for sentences labeled by annotators as Argentinian, Spanish, and both.
A likelihood over the threshold of 0.5 results in the sentence being assigned that label.
Single-label samples receive high likelihoods while multi-label samples exhibit broader distributions, reflecting mixed dialectal cues.
This behavior is consistent with our goals of encoding dialect-ness in a continuous manner to allow for nuance in analysis.

On Arabic single-label DID, \textsc{IdioleX} achieves an accuracy of 0.66, outperforming \citet{samih-etal-2019-qc} by 7 points (0.59) and coming within 1 point of \citet{abu-kwaik-saad-2019-arbdialectid} (0.67), the best reported result from the shared task. \citet{samih-etal-2019-qc}'s system, although neural, incorporates explicit heuristics such as named regional entities, whereas \textsc{IdioleX} relies purely on learned representations. In contrast, \citet{abu-kwaik-saad-2019-arbdialectid} uses a non-neural, feature-engineered approach tailored specifically to this task, which likely contributes to its strong performance. Despite this, \textsc{IdioleX}, as a general-purpose representation framework evaluated across multiple tasks, achieves near state-of-the-art performance without task-specific engineering, highlighting the strength and generality of its learned representations.

\vspace{-0.2cm}

\subsection{Authorship Attribution}

\begin{wraptable}[12]{r}{0.45\textwidth}
\small
\centering
\vspace*{-2ex}
\begin{tabular}{lc}
\toprule
\textbf{Models} & \textbf{Accuracy} \\
\midrule
\textsc{IdioleX} & 31\% \\
Fine-tuned \textsc{IdioleX} & 36\%\\
Fine-tuned \textsc{IdioleX} + Lexical & \textbf{38\%}\\
\midrule
Fine-tuned BERT & 28\% \\
Centroid Clustering w/ BERT &  16\%\\
Centroid Clustering w/ E5 &  27\%\\ 
\bottomrule
\end{tabular}
\caption{Open-set authorship attribution exact match accuracy on the PAN 19 benchmark.}
\label{tab:authattr}
\end{wraptable}

AA is another task that relies on stylistic patterns rather than semantic content.
To assess stylistic variation independent of dialect, we include the PAN 2019 Spanish authorship attribution dataset \citep{kestemont_2019_3530313}, an open-set benchmark. 
This benchmark specifically examines cross-domain capabilities: the test set data is all in a single domain that is not present during training.
The original Spanish PAN 19 shared task was conducted on the story level, so we omit submissions as a baseline since we are predicting on a sentence-level.
Sentence-level attribution is particularly challenging due to semantic constraints on feature realization and feature sparsity in short texts.

Since there are \texttt{<UNK>} author labels, we assign the \texttt{<UNK>} class label to any samples where the difference in likelihood between the top two labels is below a threshold, which was calculated per-problem via hyperparameter search over a development set (see Appendix~\ref{app:class}).

Table~\ref{tab:authattr} reports that \textsc{IdioleX} models in all training formats outperform baselines, achieving 10\% absolute improvement over the strongest comparison model.
Since PAN 19 is a cross-domain benchmark, these results demonstrate that \textsc{IdioleX} captures stylistic linguistic features that generalize beyond semantic and topical domains.

\section{Disentangling Idiolect from Semantics}
\label{sec:repval}

To evaluate whether the learned embeddings reduce semantic interference, per Section~\ref{sec:tasksem}, we examine their relationship to traditional semantic representations via comparison to idiolectal similarity scores computed by calculating the cosine similarity of representations.


\subsection{Relationship to Semantic Similarity}

\begin{figure}
\centering
\begin{minipage}[t]{0.5\linewidth}
    \centering
    \includegraphics[width=\linewidth]{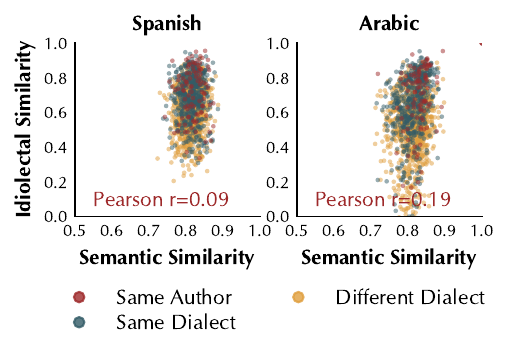}
    \caption{\textsc{IdioleX}-based stylistic similarity scores of a random sample of the withheld test set of the Reddit corpus plotted against Multilingual-E5 semantic similarity scores \citep{wang2024multilingual}. We see a low, but positive correlation.}
    \label{fig:sem_vs_style}
\end{minipage}
\hfill
\begin{minipage}[t]{0.45\linewidth}
    \centering
\includegraphics[width=0.98\linewidth]{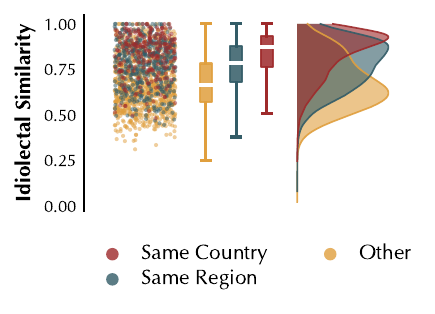}
    \caption{Stylistic similarity on the MADAR-26 dataset calculated via the Arabic \textsc{IdioleX} model on sentences with identical semantic content. We see clear differentiation by dialectal proximity.}
    \label{fig:same_sem_emb}
\end{minipage}
\end{figure}

To assess the degree to which \textsc{IdioleX} representations are influenced by semantic content, we compare the idiolectal similarity scores with semantic similarity scores computed using representations generated by Multilingual-E5 \citep{wang2024multilingual}, one of the highest performing open-source semantic embedding models according to the Massive Text Embedding Benchmark (MTEB) \citep{muennighoff-etal-2023-mteb}.

In Figure~\ref{fig:sem_vs_style}, we plot stylistic similarity against semantic similarity for a random sample of sentence pairs from the Reddit test set in both Spanish and Arabic. 
Each point corresponds to a pair of sentences, colored by relational category (same author, same dialect, different dialect).
For both languages, the correlation between semantic and idiolectal similarity is positive but weak (Pearson $\rho = 0.09$ for Spanish and $\rho = 0.19$ for Arabic).
This indicates that while stylistic similarity is not fully independent of semantic similarity, the two are only loosely coupled.
This is expected in any true idiolectal representation because some linguistic features are tied to semantic content (e.g., double negation is dependent the presence of a negative).
Still, this is not usually the case, hence the low magnitude of correlation.

\subsection{Controlling for Semantics}
To more directly test whether \textsc{IdioleX} models capture dialectal variation independent of meaning, we utilize the MADAR 26 dataset \citep{bouamor-etal-2018-madar}, which contains parallel sentences across dialects.
Because the sentences are semantically equivalent translations, any differences in similarity must arise from linguistic form rather than content.

In Figure~\ref{fig:same_sem_emb}, we plot idiolectal similarity scores on the MADAR 26 dataset, calculated on parallel samples, grouped by geographical proximity: same country (but different city), same dialectal region, or different dialectal region.
For this analysis, MSA is considered to exist in its own dialectal region.

We observe a clear gradient where geographic proximity indicates higher average stylistic similarity.
The differences between groups are statistically significant ($p < 0.005$).
Because semantic content is held constant, this pattern provides strong evidence that the \textsc{IdioleX} framework results in models that encode dialectal structure in a content-independent manner, to the point of capturing fine-grained geographic variations.

We do not claim complete disentanglement of style and semantics; as discussed in Section~\ref{sec:tasksem}, linguistic structure and semantic content depend on one another to various degrees. 
Rather, our objective is a shift in emphasis: \textsc{IdioleX} representations prioritize linguistic patterns over topicality. 
The empirical evidence suggests that this shift is achieved in practice.

\section{Applications in LLM Post-Training}
\label{sec:post}

\begin{table}[]
\small
\begin{tabular}{l|ccccc|ccccc}
\toprule
\multirow{2}{*}{\textbf{Model}} & \multicolumn{5}{c}{\textbf{ADI2}} & \multicolumn{5}{c}{\textbf{ChrF++}} \\
 & \textbf{mar} & \textbf{egy} & \textbf{pse} & \textbf{syr} & \textbf{sau} & \textbf{mar} & \textbf{egy} & \textbf{pse} & \textbf{syr} & \textbf{sau} \\
 \midrule
Allam + \textsc{Idiolex} SFT & 0.54 & \textbf{0.48} & \textbf{0.14} & 0.12 & \textbf{0.11} & \textbf{33.2} & \textbf{36.4} & \textbf{36.1} & 34.6 & 36.7 \\
Allam + SFT & 0.49 & 0.45 & 0.12 & \textbf{0.13} & \textbf{0.11} & 31.8 & 33.3 & 34.2 & \textbf{35.1} & \textbf{37.3} \\
Allam & 0.26 & 0.28 & 0.05 & 0.07 & 0.04 & 30.7 & 33.3 & 34.3 & 33.5 & 35.4 \\
\midrule
Llama 3.1+ \textsc{IdioleX} SFT & \textbf{0.60} & 0.29 & 0.12 & 0.10 & 0.07 & 27.0 & 25.4 & 27.1 & 26.4 & 29.5 \\
Llama 3.1 + SFT & 0.57 & 0.27 & 0.11 & 0.08 & 0.05 & 22.0 & 22.4 & 25.4 & 23.9 & 28.2 \\
Llama 3.1 & 0.09 & 0.12 & 0.02 & 0.02 & 0.01 & 21.1 & 20.2 & 23.9 & 21.0 & 29.0 \\
\midrule
\cite{mbzuai-citation} & 0.57 & 0.45 & 0.10 & 0.18 & {\ul 0.10} & 29.2 & {\ul 39.5} & {\ul 38.4} & {\ul 33.2} & {\ul 42.9} \\
\cite{nus-ids-citation} & {\ul 0.70} & {\ul 0.63} & {\ul 0.12} & 0.10 & 0.07 & 4.6 & 5.5 & 5.8 & 6.0 & 7.4 \\
\cite{aladdin-citation} & 0.38 & 0.15 & 0.07 & {\ul 0.22} & 0.05 & {\ul 34.5} & 31.0 & 38.3 & 17.7 & 38.2 \\
\bottomrule
\end{tabular}
\caption{\small Dialectalness (ADI2) and translation performance (ChrF++) scores on the AMIYA shared task evaluation set for Moroccan (\textbf{mar}), Egyptian (\textbf{egy}), Palestinian (\textbf{pse}), Syrian (\textbf{syr}), and Saudi (\textbf{sau}) Arabic. Our models, particularly Allam + \textsc{Idiolex} SFT, balance style and translation quality across dialects, as compared to reported scores for top submissions for the AMIYA shared task \citep{robinson-etal-2026-amiya} (bottom three rows). Best scores from our experiments are bolded, while best reported scores from the AMIYA shared task are underlined.}
\label{tab:sft}
\vspace{-0.4cm}
\end{table}

Because \textsc{IdioleX} provides a continuous measure of idiolectal similarity, it offers a possible objective for aligning LLM outputs when sensitivity to stylistic and dialectal variation is desired.
As a proof of concept, we incorporate idiolectal representations into LLM post-training with two instruction-tuned LLMs: Llama 3.1 8B,\footnote{\url{https://huggingface.co/meta-llama/Llama-3.1-8B-Instruct}} which is multilingual \citep{grattafiori2024llama}, and Allam 7B,\footnote{\url{https://huggingface.co/humain-ai/ALLaM-7B-Instruct-preview}} which is pretrained on English and Arabic only \citep{bariallam}.
We show that this leads to improved dialectal alignment without sacrificing fluency on the AMIYA shared task test set, using ADI2 dialectal adherence via the Al-Qasida evaluation framework \citep{robinson-etal-2025-al} and English $\to$ Dialectal Arabic (DA) translation quality (ChrF++) on MADAR.

\subsection{Supervised Fine-Tuning with \textsc{IdioleX}}
Standard supervised fine-tuning (SFT) optimizes cross-entropy loss, which does not explicitly encourage dialectal fidelity.
As a result, models trained with standard SFT may produce fluent outputs that lack strong dialectal alignment.
We find that with standard SFT, models that demonstrate the highest dialectal alignment (\cite{nus-ids-citation} in Table~\ref{tab:sft}) exhibit substantial degradation in fluency, as measured by ChrF++, highlighting a trade-off between alignment and generation quality.

To address this, we augment SFT with an \textsc{IdioleX}-based embedding alignment objective.
For each training sample, we precompute the \textsc{IdioleX} embedding $\mathbf{e}_i$ of the ground-truth response.
During training, we pool the LLM's hidden states $\bar{\mathbf{h}}_i$ over the generated tokens and project them into the \textsc{IdioleX} embedding space via a learned projection head $g_\theta$. The resulting loss then combines standard cross-entropy with a cosine similarity term:

\vspace{-0.2cm}
$$\mathcal{L} = \mathcal{L}_{\text{CE}} + \alpha \left(1 - \cos(g_\theta(\bar{\mathbf{h}}_i), \mathbf{e}_i)\right)$$

where $\alpha=0.5$.
Intuitively, this ensures that the same idiolectal information encoded by \textsc{IdioleX} representations is recoverable from the hidden state of the LLM.
Following \cite{mbzuai-citation}, we construct instruction–response pairs by augmenting monolingual DA text from the AMIYA closed-data track using GPT 5o Mini\footnote{\texttt{gpt-5-mini-2025-08-07}} to form instruction-answer pairs \cite{robinson-etal-2026-amiya}.
Unlike prior work, we train on a relatively small dataset (approximately 50k pairs).
Additional training details, including information on training data, are provided in Appendix~\ref{app:post}.

\subsection{Alignment Benefits}

Across all models in Table~\ref{tab:sft}, using SFT with \textsc{Idiolex} consistently improves dialectal alignment while maintaining or improving translation quality. This effect is particularly strong for Moroccan, Egyptian, and Palestinian Arabic. For example, \cite{nus-ids-citation} achieves high ADI2 scores but suffers from low ChrF++, while \cite{aladdin-citation} attains better translation performance at the cost of weaker alignment. In contrast, Allam 7B fine-tuned with \textsc{IdioleX} achieves both high dialectal alignment and competitive translation performance, despite being trained on fewer than 50k samples compared to the 2M used by \cite{aladdin-citation}.

Performance varies with data availability. Moroccan and Egyptian Arabic have the largest training sets (10,612 and 8,317 monolingual samples, respectively), and correspondingly show the strongest gains. Palestinian Arabic, by contrast, has very limited data (743 samples), both in our setup and in the shared task. Prior work (\cite{mbzuai-citation}) mitigates this by incorporating Jordanian data, which is dialectally similar. Although we also include Jordanian samples (7,217) without explicitly labeling them as Palestinian, our model still achieves strong performance on Palestinian. Our strong performance in Palestinian across metrics demonstrates that our model was able to align well dialectally to Palestinian despite the data scarcity -- 
suggesting that \textsc{IdioleX} benefits more from coverage of related idiolects and dialects than models that are trained with discrete dialect labels.

Finally, unlike AMIYA submissions, which may fine-tune separate models for each dialect, our models are trained once without dialect-specific specialization. Our strong performance across dialects indicates that \textsc{IdioleX} with SFT not only improves dialectal alignment, but also enables models to adapt their output dialect appropriately based on the input.

\section{Related Work}
Idiolectal and stylistic variation has been studied across authorship attribution, dialect identification, stylistic representation learning, and personalized language modeling. 
Our work builds on these lines by learning continuous, reusable sentence representations that capture stylistic and dialectal variation across tasks.

\paragraph{Authorship Attribution \& Dialect Identification}
Authorship attribution (AA) models individual writing style through supervised or contrastive objectives for applications such as forensic linguistics \citep{stamatatos-etal-1999-automatic, 10.1145/3339252.3340508, bischoff2020importancesuppressingdomainstyle, fabien-etal-2020-bertaa, rivera-soto-etal-2021-learning, huertastato2025partpretrainedauthorshiprepresentation}, while dialect identification (DID) focuses on linguistic variation across regions and communities for uses in text classification \citep{tiedemann-ljubesic-2012-efficient, lui-cook-2013-classifying, bouamor-etal-2018-madar}.
Both tasks rely on an overlapping set of stylistic and linguistic signals and are typically formulated as discrete classification problems, but are rarely approached in conjunction with one another.
In contrast, we leverage authorship and dialect structure as weak supervision to learn \emph{continuous representations} rather than closed-set predictions.

\paragraph{Style  Representations}
Standard sentence embeddings are optimized primarily for semantic similarity and have limited sensitivity to stylistic variation \citep{reimers2019sentence, krasnowska-kieras-wroblewska-2019-empirical, wang2024multilingual}.
Recent work has proposed learning style-aware representations, including CISR \citep{wegmann-etal-2022-author}, StyleDistance \citep{patel-etal-2025-styledistance}, and approaches that incorporate explicit linguistic features such as LISA vectors \citep{patel-etal-2023-learning}.
Our approach is closely related, but differs by jointly leveraging proximity scores gleaned from authorship and community metadata and feature prediction to learn representations that reflect stylistic similarity.

\paragraph{Personalization and Stylistic Alignment}
Personalized language modeling and user representations capture individual variation in language use, often entangling stylistic and semantic preferences \citep{lin2017personalized, welch-etal-2020-compositional, kocon2021learning}.
More recent work explores controlling the stylistic properties of LLM outputs and evaluating dialectal fidelity \citep{durandard-etal-2025-language, durandard-etal-2025-llms, robinson-etal-2025-al}.
We continue this line of work by using learned stylistic representations as training signals during post-training, enabling alignment with user-level linguistic variation.

\section{Conclusion}

We introduce \textsc{IdioleX}, a framework for idiolectal representation learning that models stylistic and dialectal variation in a continuous manner by combining supervision from proximity scores with LLM-extracted linguistic features.
We show that these representations capture meaningful idiolectal structure, generalize across domains, and support multiple downstream uses like DID and AA.
They exhibit low correlation with semantic similarity and provide an effective training objective for improving dialectal alignment in post-training.
More broadly, \textsc{IdioleX} highlights the value of treating idiolect as a graded and compositional phenomenon through a shift from discrete labels and tags to continuous representations.
This approach provides a practical and extensible framework for studying variation across languages and building tools that are more sensitive to variation.

\section*{Ethical Considerations}
All experiments in this work are conducted using existing, publicly available datasets and do not involve direct interaction with human subjects. 
Datasets derived from online sources such as Reddit may contain informal or inconsistently labeled dialectal data, without expert linguistic validation.
Consequently, both models evaluated using \textsc{IdioleX} representations trained on this data may exhibit uneven performance across language varieties.

While modeling dialectal and stylistic variation can enable more inclusive and expressive language technologies, it also raises questions about appropriateness and user preference.
Speakers may not always want language models to generate text in their own idiolect, particularly in contexts where standard or neutral varieties are expected, or where incorrect or exaggerated dialectal usage could be perceived as inauthentic or undesirable. 
This work does not evaluate user attitudes toward dialectal generation, and we do not assume that improved modeling of dialectal variation necessarily implies that models should produce such variation by default.

We view these considerations as emphasizing the importance of careful evaluation, user control, and contextual sensitivity when incorporating dialectal representations into downstream systems. 
Further work is needed to better understand how such representations interact with user preferences and social norms in deployed language technologies.

\bibliography{anthology,custom}
\bibliographystyle{colm2026_conference}

\appendix
\section{Training Data}
\label{app:data}

\begin{table*}[]
\small
\centering
\begin{tabular}{lll|lllll}
\toprule
\multirow{2}{*}{\textbf{Language}} & \multirow{2}{*}{\textbf{Subreddit}} & \multirow{2}{*}{\textbf{Dialect Label}} & \multicolumn{1}{l}{\multirow{2}{*}{\textbf{\# Authors}}} & \multicolumn{4}{l}{\textbf{\# Sentences}} \\
 &  &  & \multicolumn{1}{l}{} & \multicolumn{1}{l}{\textbf{Pre-Train}} & \multicolumn{1}{l}{\textbf{Train}} & \multicolumn{1}{l}{\textbf{Dev}} & \multicolumn{1}{l}{\textbf{Test}} \\
 \midrule
 \midrule
\multirow{16}{*}{\textbf{Arabic}} & r/algeria & Algerian & 3728 & 24003 & 5977 & 149 & 182 \\
 & r/Egypt & Egyptian & 21117 & 536429 & 16188 & 209 & 129 \\
 & r/Iraq & Iraqi & 2542 & 21595 & 7810 & 90 & 224 \\
 & r/jordan & Jordanian & 14745 & 427230 & 15159 & 272 & 204 \\
 & r/Kuwait & Kuwaiti & 1739 & 13099 & 6027 & 88 & 228 \\
 & r/lebanon & Lebanese & 1705 & 12020 & 7083 & 413 & 227 \\
 & r/Libya & Libyan & 788 & 6344 & 6202 & 298 & 238 \\
 & r/Morocco & Moroccan & 3462 & 27940 & 6131 & 121 & 243 \\
 & r/Oman & Omani & 516 & 1515 & 1515 & 106 & 102 \\
 & r/Palestine & Palestinian & 530 & 2016 & 2016 & 151 & 252 \\
 & r/qatar & Qatari & 402 & 794 & 794 & 93 & 119 \\
 & r/saudiarabia & Saudi & 25982 & 639810 & 10743 & 201 & 463 \\
 & r/Sudan & Sudanese & 795 & 9297 & 9297 & 327 & 125 \\
 & r/Syria & Syrian & 1765 & 48265 & 13355 & 598 & 394 \\
 & r/UAE & Emirati & 318 & 458 & 458 & 93 & 98 \\
 \midrule
 & \textbf{Total} &  & \textbf{80134} & \textbf{1770815} & \textbf{108755} & \textbf{3209} & \textbf{3228} \\
 \midrule
 \midrule
\multirow{18}{*}{\textbf{Spanish}} & r/argentina & Argentinian & 122963 & 21532596 & 41160 & 399 & 4607 \\
 & r/BOLIVIA & Bolivian & 6431 & 196137 & 12431 & 401 & 742 \\
 & r/chile & Chilean & 65314 & 7563763 & 35592 & 850 & 1819 \\
 & r/Colombia & Colombian & 56421 & 2166716 & 19553 & 1246 & 159 \\
 & r/cuba & Cuban & 2846 & 65180 & 9191 & 529 & 1458 \\
 & r/Dominican & Dominican & 3892 & 65488 & 6966 & 455 & 105 \\
 & r/ecuador & Ecuadorian & 5628 & 163765 & 9003 & 145 & 244 \\
 & r/ElSalvador & El Salvadorian & 4119 & 154321 & 13914 & 91 & 189 \\
 & r/guatemala & Guatemalan & 10756 & 404417 & 15783 & 1299 & 621 \\
 & r/Honduras & Honduran & 6882 & 281450 & 19740 & 364 & 1989 \\
 & r/mexico & Mexican & 151978 & 8729427 & 50212 & 792 & 1500 \\
 & r/Panama & Panamanian & 13608 & 1016097 & 17441 & 459 & 950 \\
 & r/Paraguay & Paraguayan & 9845 & 479757 & 15455 & 675 & 225 \\
 & r/PERU & Peruvian & 23411 & 792837 & 12280 & 2054 & 431 \\
 & r/spain & Peninsular & 18903 & 395424 & 7792 & 262 & 384 \\
 & r/uruguay & Uruguayan & 29093 & 3274405 & 33568 & 784 & 1694 \\
 & r/vzla & Venezuelan & 25078 & 2202730 & 16640 & 481 & 2120 \\
 \midrule
 & \textbf{Total} &  & \textbf{557168} & \textbf{49484510} & \textbf{336721} & \textbf{11286} & \textbf{19237} \\
 \bottomrule
\end{tabular}
\caption{Data split sizes for Reddit data by dialect.}
\label{tab:fulldata}
\end{table*}

\subsection{Sampling}

We collect publicly available Reddit comments using the Pushshift archive for each subreddit listed in Table~\ref{tab:fulldata}, spanning from each subreddit's inception through December 2024 \citep{baumgartner2020pushshift}.
Comments are segmented into sentences by punctuation using regex, and language filtered using fastText language identification. Only sentences classified as the target language are retained.

Users that appear as \texttt{[deleted]}, \texttt{[removed]}, and \texttt{AutoModerator} are thrown out since they do not belong to one consistent user.
Any sentences with fewer than 5 words (under the definition of \textit{word} as a space-delineated token) are removed and any comments with fewer than 2 sentences or users with fewer than 2 comments are removed due to constraints around batch construction.

\subsection{Splitting}

From the original set of authors for each dialect, 10 are randomly selected to be in each of the dev and test splits.
This ensures that the dev and test splits are made up of completely unseen authors to avoid contamination with the train set.

Of the remaining authors, all are present in the pre-train set. 
The train set contains a subset of the pre-train set and is the set for which feature extraction is performed.
This set is made up of 200 authors per dialect, or all of the authors in the pre-train set, whichever is lower.

\subsection{Feature Extraction}

\label{app:feat_extraction}

\begin{table*}[t]
\scriptsize
\centering
\setlength{\tabcolsep}{6pt}
\begin{tabular}{ll}
\toprule
\multicolumn{2}{c}{\textbf{Spanish Binary Features}} \\
\midrule
\multicolumn{2}{l}{\textbf{Subject Pronouns \& Verbal Morphology}} \\
\midrule
contains explicit subject yo & contains 2pl present suffix ais \\
contains explicit subject tu & contains 2pl present suffix eis \\
contains explicit subject usted & contains 2pl present suffix is \\
contains explicit subject vos & contains voseo present suffix as stressed \\
contains explicit subject vosotros or vosotras & contains voseo present suffix es stressed \\
contains explicit subject ustedes & contains voseo imperative form \\
contains explicit subject nosotros or nosotras & \\
contains explicit subject ellos or ellas & \\
\midrule
\multicolumn{2}{l}{\textbf{Diminutives}} \\
\midrule
contains diminutive suffix ito or ita & contains diminutive suffix illo or illa \\
contains diminutive suffix ico or ica & contains diminutive suffix ino or ina \\
contains diminutive suffix in regional & contains diminutive suffix ete \\
contains diminutive suffix ingo or inga & contains diminutive suffix iquio or iquia \\
contains diminutive suffix icho or icha & \\
\midrule
\multicolumn{2}{l}{\textbf{Clitic \& Orthographic Variation}} \\
\midrule
contains differential object marking a before animate direct object & contains preverbal object clitic \\
contains accusative clitic doubling lo a NP & contains enclitic on infinitive \\
contains accusative clitic doubling la a NP & contains enclitic on gerund \\
contains accusative clitic doubling los a NP & contains multiple clitics in sequence \\
contains accusative clitic doubling las a NP & contains se plus object clitic \\
contains dative clitic doubling le a NP & contains strong pronoun object doubling \\
contains dative clitic doubling les a NP & contains indirect object strong pronoun doubling \\
contains inverted question mark & contains inverted exclamation mark \\
contains all caps word & contains repeated punctuation \\
\bottomrule
\end{tabular}
\caption{Spanish linguistic features extracted via LLM for \textsc{IdioleX} training. All features are sentence-level binary indicators.}
\label{tab:spanish_feats}
\end{table*}

\begin{table*}[t]
\scriptsize
\centering
\setlength{\tabcolsep}{6pt}
\begin{tabular}{ll}
\toprule
\multicolumn{2}{c}{\textbf{Arabic Binary Features}} \\
\midrule
\multicolumn{2}{l}{\textbf{Morphosyntax \& Clause Structure}} \\
\midrule
contains case endings u a i & contains future prefix sa \\
contains tanwin un an in & contains future particle sawfa \\
contains dual suffix an or ayn & contains dialectal future marker rah ha ghadi bash \\
contains 2sg feminine suffix ki & contains progressive marker 3am qa3id jaalis ka bi \\
contains 2sg feminine dialect suffix sh or ish & contains imperfect prefix bi or ba \\
contains pronoun hum & contains imperfect prefix ka \\
contains pronoun hunna & contains imperfect prefix 3a \\
contains dialect possessive suffix o or u & contains analytic passive in it ta \\
is verb initial clause & contains internal passive kutiba \\
is subject initial clause & contains passive prefix ta it et \\
contains overt present copula & contains relative alladhi forms \\
contains zero copula equational & contains relative illi yalli elli \\
\midrule
\multicolumn{2}{l}{\textbf{Negation, Interrogatives \& Discourse Particles}} \\
\midrule
contains negator laysa & contains yes no particle hal \\
contains negator lan & contains dialect yes no particle huwwa wash ish \\
contains negator lam & contains interrogative ayy \\
contains negator lamma & contains interrogative eh \\
contains ma verbal negation & contains interrogative shu \\
contains ma verb sh negation & contains interrogative wesh \\
contains negator mish or mesh & contains interrogative shinu \\
contains negator mu or mo & contains discourse wallahi \\
contains egyptian ayywa & contains discourse ya3ni \\
contains msa adverb haqqan & \\
\midrule
\multicolumn{2}{l}{\textbf{Dialectal Lexical Markers \& Orthography}} \\
\midrule
contains egyptian izaay & contains egyptian demonstrative da di dol \\
contains egyptian keda & contains levantine demonstrative hada haydi \\
contains egyptian ba2a & contains maghrebi demonstrative had \\
contains levantine baddi & contains gulf wain \\
contains levantine halla & contains gulf negator mo \\
contains levantine shu & contains gulf yalla \\
contains levantine demonstrative hada & contains maghrebi barcha \\
contains maghrebi hshuma & contains maghrebi prefix ka \\
contains yemeni maish & contains yemeni shill \\
contains nonstandard qaf gaf & contains nonstandard hamza omission \\
contains nonstandard alef maqsura variation & contains arabic question mark\\
contains question mark & contains exclamation mark ! \\
contains repeated punctuation & contains ellipsis \\
contains tatweel & contains letter lengthening \\
contains laughter token & \\
\bottomrule
\end{tabular}
\caption{Arabic linguistic features extracted via LLM for \textsc{IdioleX} training. All features are sentence-level binary indicators.}
\label{tab:arabic_feats}
\end{table*}

\subsubsection{Feature Inventory}

We use sets of linguistic features extracted from language-specific dialect resources \citep{ArabicTextbook, SpanishTextbook}.
These features focus primarily on morphology/morphosyntax, affixes and functional lexemes, and orthography.
A full list of Spanish features can be found in Table~\ref{tab:spanish_feats} and a full list of Arabic features can be found in Table~\ref{tab:arabic_feats}.

\subsubsection{Prompting}

To annotate sentences with the linguistic features they contain, we prompt an LLM (in our case, GPT 5 mini\footnote{gpt-5-mini-2025-08-07}).
The LLM is asked to return a JSON dictionary of binary features with the following prompt:
\begin{quote}
You are a deterministic feature extractor.

Task: For each feature key, output 1 if the sentence contains an explicit surface cue described by that feature; otherwise, output 0.

Use a recall-oriented policy: if the cue is reasonably present, set 1.
Return VALID JSON ONLY with exactly one top-level key: 'features'.

'features' maps each feature string EXACTLY (copy verbatim) to 0 or 1.
    
Feature keys: {features}
Sentence: {sentence}
\end{quote}
In cases where the LLM returns an invalid response (i.e. due to safety filtering), we default to marking every feature as 0. 

\subsection{Batching}

Each batch begins with an \textbf{anchor} sentence being sampled, which acts as the seed for the rest of the batch.
This anchor sentence is randomly sampled from the dataset.

The other sentences in the batch fall into the following categories:
\begin{enumerate}
    \item \textit{Same Comment}: A different sentence from the same content. $r = 3$
    \item \textit{Same Author}: A sentence from a different comment by the author. $r = 2$
    \item \textit{Same Subreddit}: A sentence from the same subreddit/region but written by a different author. $r = 1$
    \item \textit{Different Subreddit}: A sentence from a different subreddit/region versus the anchor. $r = 0$
\end{enumerate}
This is done by sampling randomly over the set of sentences from the comment, the set of comments by the same author, the set of authors from the same subreddit, and the set of subreddits.

For training efficiency, we structure the batch such that every sentence has at least one sentence that falls into one of these categories.
In practice, to ensure this, we use mini-batches of size 16 where for each anchor $x_a$, there are $2^{3 - n}
$ samples with proximity score $n \in [0,3]$.
In other words, for every sentence, there will be 1 other sentence from the same comment, 2 from the same author but a different comment, 4 from a different author but the same subreddit, and 8 from another subreddit.

\section{Experimental Details}

\subsection{Training Under the \textsc{IdioleX} Framework}
\label{app:train_details}

\subsubsection{Base Models}

We used the the most prolific pre-trained monolingual language model for each language.
We focused on models with fewer than 200 M parameters to ensure that the resulting model would be lightweight enough to use as a training objective or reward model.
Specifically, for Spanish we use BERTIN\footnote{\url{https://huggingface.co/bertin-project/bertin-roberta-base-spanish}}, a RoBERTa-base model trained on the Spanish component of mC4 \citep{delarosa2022bertinefficientpretrainingspanish}.
For Arabic, we use AraBERT v2\footnote{\url{https://huggingface.co/aubmindlab/bert-base-arabertv2}}, a BERT-base model trained on Arabic news corpora \citep{antoun2020arabert}.
Both models follow the standard BERT-base architecture with 12 transformer layers and a hidden dimensionality of 768.

Tokenization is performed using the HuggingFace tokenizer corresponding to each pretrained model. 
Inputs are truncated to a maximum length of 512 tokens.

\subsubsection{Auxiliary Prediction Heads}

To incorporate linguistic supervision, we add two auxiliary heads on top of the encoder.

\paragraph{Feature Prediction}
A two-layer feedforward network predicts the binary feature vector associated with each sentence representation. 
The architecture consists of:
\[
\text{Linear}(d,2d) \rightarrow \text{ReLU} \rightarrow \text{Linear}(2d,F)
\]
where $d$ is the encoder hidden size and $F$ is the number of linguistic features (different per language).
The model is trained using binary cross-entropy with logits.

\paragraph{Contrastive Similarity Represenations}
For supervised contrastive learning, representations are passed through a projection head:
\[
\text{Linear}(d,d) \rightarrow \text{ReLU} \rightarrow \text{Linear}(d,256).
\]
The resulting vectors are $\ell_2$ normalized before computing contrastive similarity based on the Jaccard similarity of the feature vectors.

\subsubsection{Training Objectives}

Training combines three losses.

\paragraph{Margin Ranking Loss (MRL)}
The margin ranking loss encourages sentences with higher proximity score to be closer in embedding space.
The margin $\lambda$ is linearly increased from $0$ to $0.5$ during training.

\paragraph{Supervised Contrastive Loss (SCL)}
We additionally apply a supervised contrastive loss where the similarity between samples is weighted using the Jaccard similarity between their feature vectors. 

\paragraph{Feature prediction loss.}
The feature prediction head is trained using binary cross-entropy on the predicted feature logits.

\paragraph{Final Objective}
The final training objective is
\begin{equation}
L = (1-\alpha)L_{rank} + \alpha L_{feat},
\end{equation}
where
\[
L_{feat} = 0.25L_{BCE} + L_{supcon}.
\]

\subsubsection{Variance and Decorrelation Regularization}

To prevent representation collapse and reduce embedding anisotropy, we apply a VICReg-style regularization term consisting of:

\begin{itemize}
\item a variance constraint encouraging the standard deviation of each embedding dimension to exceed 1,
\item a decorrelation penalty on off-diagonal covariance entries.
\end{itemize}

\subsubsection{Sentence Representation Construction}

Sentence embeddings are obtained from the transformer using a layer-wise attention mechanism similar to that used by \textsc{COMET} \citep{rei2020comet}. 
Let $h_l$ denote the hidden states from layer $l$. 
A learned scalar weight $w_l$ is applied to each layer and normalized using a softmax:

\begin{equation}
\tilde{h} = \sum_{l=0}^{L} \alpha_l h_l, \quad
\alpha_l = \frac{\exp(w_l)}{\sum_{k=0}^{L} \exp(w_k)}.
\end{equation}

The resulting token representations are then aggregated using average pooling over all non-padding tokens to obtain a sentence-level representation.

During training, embeddings are mean-centered using a running estimate of the global embedding mean across all distributed workers. 
After centering, vectors are $\ell_2$ normalized.

\subsubsection{Training Procedure}

Training proceeds in two stages.

\paragraph{Stage 1: Ranking Pretraining}
The encoder is trained using only the margin ranking loss on the full dataset. 
This stage uses all available sentences.

\paragraph{Stage 2: Feature-Aware Training}
Training is then continued on the subset of sentences with feature annotations. 
During this stage the feature prediction and contrastive losses are added.

\subsubsection{Optimization}

Models are trained using the Adam optimizer with a learning rate of $1\times10^{-5}$ and linear warmup for 25k steps.

Training uses a batch size of 32 sentences. 
Supervised contrastive learning uses a temperature $\tau=0.07$, and Jaccard weights are truncated to the top 5 neighbors.

\subsubsection{Training Infrastructure}

Training is performed using PyTorch DistributedDataParallel across 4 GPUs. 
All models are implemented in PyTorch and trained using the HuggingFace Transformers library.

\subsubsection{Training Hyperparameters}

\label{app:hyperparameters}

Table~\ref{tab:training_hyperparameters} summarizes the primary hyperparameters used for training using the \textsc{IdioleX} framework. 
Unless otherwise specified, the same configuration is used across all languages, with only the underlying pretrained encoder differing.

\begin{table}[h]
\centering
\small
\begin{tabular}{ll}
\toprule
\textbf{Parameter} & \textbf{Value} \\
\midrule
Hidden size & 768 \\
Transformer layers & 12 \\
Maximum sequence length & 512 \\

\midrule
Batch size & 32 \\
Ranking group size & 16 \\
Pretraining epochs & 3 \\
Full training epochs & 10 \\
Validation frequency & 250 steps \\

\midrule
Margin ranking margin & $0 \rightarrow 0.5$\\
Feature loss weight ($\alpha$) & 0.5 \\
Feature BCE weight & 0.25 \\
Supervised contrastive temperature $\tau$ & 0.07 \\
Jaccard top-$k$ positives & 5 \\

\midrule
Feature head hidden size & $2d$ \\
Projection head output size & 256 \\
Feature vector dimension & 41 \\

\midrule
Learning rate & $1 \times 10^{-5}$ \\
Learning rate warmup & 25k steps \\
Early stopping patience & 25 \\

\midrule
GPUs & 4 \\
Hours & $\leq 48$ hrs \\

\bottomrule
\end{tabular}
\caption{Training hyperparameters used for \textsc{IdioleX}.}
\label{tab:training_hyperparameters}
\end{table}

\subsection{Evaluation Protocols}

\subsubsection{\textsc{IdioleX}-based Classification Procedure}
\label{app:class}

To evaluate whether models trained via the \textsc{IdioleX} framework can support downstream classification tasks, we fine-tune classification heads on top of the pretrained encoder models.
This section describes the training procedure, optimization settings, and implementation details used to produce the classification results reported in Section~\ref{sec:class}.

\paragraph{Models} For each classification task, we initialize a sequence classification model using the same pretrained encoder architecture used in the representation learning stage; that is, AraBERT v2 \citep{antoun2020arabert} for Arabic and BERTIN \citep{delarosa2022bertinefficientpretrainingspanish} for Spanish. 
The encoder weights are loaded directly from the \textsc{IdioleX} checkpoint containing pretrained encoder parameters. 
The \textsc{IdioleX} checkpoints contain only encoder weights; the classification layer is newly initialized to match the number of target classes. 

The classification head consists of a linear projection from the encoder’s pooled representation to the label space. 
For multi-label tasks (DSL-ML), the model is configured for multi-label classification, whereas single-label (MADAR-26, PAN-19) tasks use standard softmax classification.

\paragraph{Training} Training is performed using the AdamW optimizer with linear learning rate decay and warmup. The learning rate schedule follows a linear warmup followed by linear decay over the course of training steps. 
Unless otherwise stated, the training hyperparameters found in Table~\ref{tab:hyp_class} are used.
\begin{table}[H]
    \centering
    \small
    \begin{tabular}{l|l}
        \toprule
        Hyperparameter & Value \\
        \midrule
        Learning Rate & $1 \times 10^{-5}$ \\
        Batch Size & 32 \\
        Weight Decay & 0.01 \\
        Warmup Ratio & 0.1 \\
        Early Stopping Patience & 3 epochs \\
        Max Sequence Length & 512 tokens \\
        \bottomrule
    \end{tabular}
    \caption{Hyperparameters used for classification fine-tuning.}
    \label{tab:hyp_class}
\end{table}
Gradient clipping with a maximum norm of 1.0 is applied to stabilize training. Mixed precision training (FP16) is optionally used when GPU support is available (always in our case). 
Because MADAR 26 consists of 26 classes, we apply an additional label smoothing of 0.1 to improve performance for that classifier. 
Training proceeds for a maximum of 25 epochs, with early stopping based on validation macro F1 score.

In tasks that allow unknown classes (e.g., open-set authorship attribution), an additional \texttt{<UNK>} label is assigned to examples whose predicted confidence falls below a tuned threshold.
This threshold is determined via a held out set of 50\% of the validation data. 

\paragraph{Lexical Ensembling}
For some experiments, a lexical model is combined with the neural classifier, reflecting performance improvements that past work has found for specific datasets. 
The lexical component consists of a TF-IDF representation over character n-grams followed by a logistic regression classifier trained using the SAGA solver.
The lexical classifier's model settings are reflected in Table~\ref{tab:hyp_lex}.
\begin{table}[H]
    \centering
    \small
    \begin{tabular}{l|l}
        \toprule
        Hyperparameter & Value \\
        \midrule
        Learning Rate & $1 \times 10^{-5}$ \\
        Batch Size & 32 \\
        Weight Decay & 0.01 \\
        Warmup Ratio & 0.1 \\
        Early Stopping Patience & 3 epochs \\
        Max Sequence Length & 512 tokens \\
        \bottomrule
    \end{tabular}
    \caption{Hyperparameters used for the lexical component of the classifier.}
    \label{tab:hyp_lex}
\end{table}
The lexical classifier outputs probability distributions over classes, which are linearly combined with the neural model predictions:
$$P = \omega P_{\text{lex}} + (1-\omega)P_{\text{neural}}$$
The ensemble weight $\omega$ is tuned on the validation set to maximize macro F1.

\paragraph{Evaluation Metrics}
Evaluation follows the protocols described in Section~\ref{sec:class}.
Predictions are evaluated using macro-averaged F1 score and exact match accuracy (important for multi-label tasks, where it evaluates how well the model predicts \textit{all} the correct labels for each item). 

\subsubsection{Classification Baselines}
\label{app:class_baseline}

\begin{table}[t]
\centering
\small
\begin{tabular}{ll|cc|cc|c}
\toprule
\multirow{2}{*}{Language} & \multirow{2}{*}{Model} & \multicolumn{2}{c}{KMeans} & \multicolumn{2}{c}{Centroid-Based} & \multicolumn{1}{c}{Retrieval} \\
 &  & F1-macro & Accuracy & F1-macro & Accuracy & Accuracy* \\
\midrule
Arabic (MADAR 26) & BERT & 0.11 & 0.14 & 0.20 & 0.22 & 0.10 \\
 & E5 & 0.04 & 0.07 & 0.15 & 0.17 & 0.06 \\
Spanish (DSL-ML) & BERT & 0.50 & 0.46 & 0.77 & 0.57 & 0.07 \\
 & E5 & 0.44 & 0.45 & 0.74 & 0.54 & 0.06 \\
\cmidrule{1-7}
Spanish (PAN 19) & BERT & 0.07 & 0.32 & 0.14 & 0.16 & 0.07 \\
 & E5 & 0.10 & 0.33 & 0.21 & 0.27 & 0.08 \\
\bottomrule
\end{tabular}
\caption{Clustering and retrieval-based evaluation of dialectal structure in sentence embedding spaces.}
\label{tab:clustering_baseline_results}
\end{table}

\paragraph{Closed-set Classification} We fine-tune the same monolingual models used in our \textsc{IdioleX} training, BERTIN~\citep{delarosa2022bertinefficientpretrainingspanish} for Spanish tasks and AraBERTv2~\citep{antoun2020arabert} for Arabic tasks, on the three classification datasets used in our experiments: MADAR 26 (Arabic), DSL-ML (Spanish), and PAN 19 (Spanish). To maintain consistency with \textsc{IdioleX}'s representation-based evaluation, we use the same logistic regression probe as in the \textsc{IdioleX} classification setup. Specifically, we extract the [CLS] representations from the fine-tuned baseline models and use them as input features to the logistic regression probe; these constitute the reported baseline results.

\paragraph{Clustering} To evaluate how well embedding models capture dialect-specific structure without supervised classification, we employ a set of clustering-based baselines over sentence-level embeddings. We conduct all experiments using BERTIN~\citep{delarosa2022bertinefficientpretrainingspanish} for Spanish and AraBERTv2~\citep{antoun2020arabert} for Arabic (the same monolingual BERT-based models used in our closed-set baseline), as well as Multilingual E5~\citep{wang2024multilingual}, a state-of-the-art multilingual sentence-embedding model commonly used in recent embedding evaluation work. Using frozen encoder representations, we fit KMeans on the training-set embeddings and then apply the fitted model to the test-set embeddings, assessing the resulting structure using \emph{F1} and \emph{Accuracy}.

Our Clustering-Based evaluation framework consists of three complementary metric categories, each capturing different aspects of dialect separation in the embedding space:

\textit{Centroid-Based Classification.} We compute class centroids by averaging embeddings from the training set for each dialect. We then classify test samples by computing the Euclidean distance from each test embedding to all training-set centroids and assigning each test sample to the nearest centroid. This provides a simple nearest-centroid baseline that evaluates how well embeddings support geometric separation of dialects. F1-macro and accuracy are computed by comparing the predicted dialect labels (from the nearest centroid) to the true dialect labels for each test sample. We report both F1-macro and accuracy, where higher values indicate better dialect separation in the embedding space.

\textit{KMeans Clustering Metrics.} We fit KMeans clustering on the training-set embeddings, using the number of unique dialect labels as the number of clusters. We then apply the fitted model to the test-set embeddings and evaluate the clustering quality using F1-macro and accuracy, which measure how well the unsupervised clusters align with the true dialect labels. F1-macro and accuracy are computed by mapping each cluster to its dominant dialect label (the most frequent true label in that cluster) and then comparing the predicted cluster assignments to the true dialect labels. Higher values indicate that the embedding space naturally separates dialects without explicit supervision.

\textit{Retrieval Baseline.} For each anchor sentence, we sample 3 positive examples (same dialect) and 3 negative examples (different dialects), then rank all 6 candidates by cosine similarity to the anchor. We report the fraction of trials where all positives rank above all negatives (\emph{Accuracy*}), which measures the model's ability to distinguish same-dialect from different-dialect samples in a retrieval setting.

Tables~\ref{tab:clustering_baseline_results} presents comprehensive results across all metrics and datasets. The tables show that while both models capture some dialect structure, there is significant variation across languages and metric types, with E5 generally performing better on centroid-based metrics while BERTIN shows competitive performance on retrieval tasks.

\subsubsection{Post-Training}
\label{app:post}

We incorporate \textsc{IdioleX} into the supervised fine-tuning (SFT) stage of post-training by integrating it into the training objective.
By operating on the model's internal representations rather than on the generated text, we avoid issues stemming from length mismatches between ground truth (or model length limits) and generations.

\paragraph{Data}
 
The original AMIYA training data is primarily monolingual text data, not instruction-tuning pairs.
To ensure our fine-tuning didn't degrade instruction-following abilities, we use a method similar to that by \citet{mbzuai-citation} to convert these corpora into instruction-response pairs, using a combination of template-based methods and LLM-based augmentation.

We draw from 9 corpora spanning 5 target dialects (Egyptian, Moroccan, Palestinian, Saudi, Syrian) as well as 5 additional dialects (Jordanian, Iraqi, Yemeni, Tunisian, and Kuwaiti):

\begin{table}[h]
    \scriptsize
    \centering
    \begin{tabular}{llllll}
        \toprule
        Source  & Type & Dialects & Size & Method \\
        \midrule
        MADAR-26 (train) \citep{bouamor-etal-2018-madar} & Bitext & Multi-Dialectal & 2k sentences per & Template \\
        FLORES-200 \citep{goyal-etal-2022-flores} & Bitext & Multi-Dialectal & 1k sentences per & Template \\
        SauDial \citep{alanazi2025saudial} & Bitext & Saudi & 1k sentences & Template \\
        \midrule
        PALM (train) \citep{alwajih-etal-2025-palm} & Monolingual & Multi-Dialectal & 4k sentences & LLM \\
        HABIBI \citep{xel-haj-2020-habibi} &  Monolingual & Multi-Dialectal & 525k verses & LLM \\
        EDC \citep{tarmom2020compression} &  Monolingual & Egyptian & 13k sentences & LLM \\
        Atlaset \citep{bounhar2025atlaset} & Monolingual & Moroccan & 1.1M tokens & LLM \\
        MASC \citep{al2016arabic} & Monolingual & Multi-Dialectal & 6k sentences & LLM \\
        JODA \cite{abandah2025jordanian} & Monolingual & Jordanian & 50k sentences & LLM \\
        \bottomrule
    \end{tabular}
    \caption{\small AMIYA data used for post-training and how it was augmented to make instruction-following data.}
    \label{tab:post_data}
\end{table}

For parallel corpora containing aligned translations, we create instruction pairs without any LLM calls. Each source sentence is prefixed with a randomly sampled translation instruction template:
\begin{enumerate}
    \item English $\to$ DA: "Translate this to [dialect]: [source]", "Say this in [dialect]: [source]", "How would you say this in [dialect]? [source]"
    \item MSA $\to$ DA: "Convert this to [dialect]: [source]", "Rewrite this in [dialect]: [source]"
\end{enumerate}

The dialectal translation serves as the ground-truth response. This yields high-quality, naturally dialectal training pairs.

For monolingual corpora (containing only dialectal text without paired translations), we use a two-step LLM augmentation process to generate full instruction-response pairs. 
First, for each dialectal corpus text, we prompt GPT-5-mini to generate a dialectal question that the text could naturally answer.
\begin{quote}
You are a native speaker of [dialect]. 
Below is a text in [dialect]. Write a short question (1 sentence) in [dialect] that someone might ask a chatbot, where this text would be a natural answer.
Rules: Write ONLY the question. 
The question must be in [dialect], NOT Modern Standard Arabic. 
Use dialectal question words like: [dialect-specific examples]. The question should be about [randomly sampled topic].
\end{quote}

Topics are sampled from 8 categories aligned with the AL-QASIDA evaluation: food/recipes, proverbs, cultural traditions, geography, history, dialect expressions, nature, and arts. Dialect-specific question words are provided to encourage authentic dialectal phrasing.

A naive approach would use the original corpus text directly as the ground-truth response. 
However, we found this produced poor SFT results: corpus fragments are typically 3-5 words (e.g., song lyrics, social media posts), and models trained on these learned to produce extremely short, often incoherent responses.
We instead prompt the LLM to generate a full conversational answer incorporating the corpus text:

\begin{quote}
You are a helpful chatbot that ONLY speaks [dialect]. A user asked you: [generated question]
Write a helpful, detailed answer (3-6 sentences) in [dialect]. You MUST incorporate this phrase naturally in your answer: [corpus text]
\end{quote}

This produces training pairs where the question is dialectal, the answer is a full multi-sentence response in dialect, and the original human-written dialectal text is preserved within the answer, ensuring the ground truth contains authentic dialectal vocabulary and phrasing rather than just LLM-generated approximations.

Due to computational constraints, we randomly sample at most 5k sentences per dataset for our monolingual training data, resulting in about 39k instruction-answer pairs total plus 10k translation pairs.

\paragraph{Architecture} 
Our approach adds two components to a standard LoRA-based SFT.
First, we use an encoder trained under the \textsc{IdioleX} framework on the language of interest. 
The parameters of this model stay frozen during SFT.
To reduce per-step overhead, we pre-compute the \textsc{IdioleX} embeddings of the ground truth data and store it in a cache.
Then, we introduce a small two-layer projection head $g_\theta$ that maps from the LLM's hidden dimension to the \textsc{IdioleX} embedding dimension.
The output of this projection head is L2-normalized in the same manner as the \textsc{IdioleX} representations are to ensure they fall in the same normalized embedding space.

For each training sample $i$ with input tokens $\mathbf{x}_i$ and response starting at position $s_i$, the combined loss is:
 
$$\mathcal{L}_i = \mathcal{L}_{\text{CE}}(\mathbf{x}_i, s_i) + \alpha \cdot \mathcal{L}_{\text{A}}(\mathbf{x}_i, s_i, \mathbf{e}_i)$$

The cross-entropy loss ($\mathcal{L}_{\text{CE}}$) is computed only over response tokens. 
The alignment loss ($\mathcal{L}_{\text{A}}$) pools the LLM's last-layer hidden states over response tokens and projects to IdioleX space:
 
$$\mathcal{L}_{\text{emb}} = 1 - \text{cosine\_similarity}\left(g_\theta(\mathbf{h}_i), \mathbf{e}_i\right)$$
 
where $\mathbf{h}_t$ is the mean-pooled hidden state at position $t$ from the last transformer layer.

\paragraph{Experimental Details}
We use $\alpha = 0.5$, LoRA rank 32 on all linear layers, a learning rate of $2 \times 10^{-4}$ with cosine annealing, and train for 1 epoch. 
The projection head adds approximately 6M parameters (for a 4096-dimensional LLM projecting to 768-dimensional \textsc{IdioleX} embedding space).
 
\textsc{IdioleX} embeddings are precomputed once on a single GPU before distributed training begins, cached to disk, and loaded by all workers. 
The LLM with LoRA adapters and the projection head are wrapped into a single `nn.Module` for compatibility with DeepSpeed ZeRO-2, which requires a single model instance. 
For efficiency, training was done with gradient checkpointing and mixed precision (bfloat16).

\end{document}